\documentclass[11pt]{article}

\usepackage{acl}

\usepackage{times}
\usepackage{latexsym}
\usepackage[T1]{fontenc}
\usepackage[utf8]{inputenc}
\usepackage{microtype}
\usepackage{inconsolata}

\usepackage{graphicx}
\graphicspath{{figures/}}

\usepackage{booktabs}
\usepackage{array}
\usepackage{multirow}
\usepackage{placeins}  % \FloatBarrier to flush floats before a section

\usepackage{amsmath}
\usepackage{amssymb}

\usepackage{url}
\newcommand{\m}[1]{\nolinkurl{#1}}

\title{Graph Serializations for Reasoning LLMs:\\
A Study Across Knowledge Extraction and Structural Reasoning}

\author{
  \textbf{Julien Yu},
  \textbf{Mihaela Bornea},
  \textbf{Avirup Sil},
  \textbf{Katrin Kirchhoff}
  \\
  Oracle
  \\
  \texttt{\{julien.yu, mihaela.bornea, avi.sil, katrin.kirchhoff\}@oracle.com}
}

\begin{document}
\maketitle

\begin{abstract}
The way a knowledge graph is serialized as text for inclusion into LLM context can substantially affect downstream LLM performance, with prior work reporting tens of percentage points variations across formats. However, existing studies consider only a limited set of representations and predate frontier reasoning models, which now offer stronger reasoning, instruction-following, and long-context capabilities. We evaluate 15 graph representation formats across nine frontier LLM configurations on ontology-based knowledge extraction and structural graph reasoning tasks. While knowledge extraction is relatively insensitive to different serialization formats, the choice of the optimal format for structural reasoning tasks significantly affects accuracy. We demonstrate that the optimal representation choice depends on model, query, and task and propose to approach graph serialization as a dynamic orchestration decision.
\end{abstract}

\section{Introduction}
\label{sec:introduction}

LLMs increasingly serve as reasoning components in systems that consume symbolic structure, including ontologies, knowledge graphs (KGs), and graph-derived retrieval contexts. Applications include ontology-guided extraction \citep{mihindukulasooriya2023t2kbench}, multi-hop reasoning over KGs \citep{sun2024tog,ma2025tog2,zhou2025reflection}, and generation grounded in retrieved graph evidence \citep{edge2024graphrag}. While it is possible to connect LLMs to dedicated graph query engines, it is increasingly common to inject graph representations directly into LLM context, via serialization of nodes, edges, types, relations, and constraints into a sequence of tokens. Prior work shows that the representation choice can substantially affect downstream task accuracy while holding the model and task fixed \citep{fatemi2024talklikeagraph,markowitz2025kgllmbench,sclar2024formatspread,he2025promptformat,sui2024table}.  However, previous studies only investigated a small number of possible graph representations, did not test frontier reasoning models, and did not analyze the relationship between query/task type, graph properties, models, and reasoning level. In particular, it remains unclear whether frontier reasoning models render serialization less important (by being able to infer structure from most reasonable representations), or more important by exploiting semantic, constraint, or layout cues when exposed explicitly.

%This paper studies the resulting representation selection problem: \emph{what graph representation should be placed in the context window of a %frontier reasoning model?} Existing work establishes that textualization matters, but it does not settle which representational properties %matter for modern KG-grounded LLM systems. 

We organize the study around three research questions:
%\textbf{RQ1: Semantic extraction.} For ontology-guided semantic knowledge extraction, do representations that expose types, domains, ranges, and constraint-like structure work better than compact or connectivity-only representations? 
%\textbf{RQ2: Structural reasoning.} For queries that crucially depend on correct analysis of graph structure (adjacency, paths, degrees, connectivity, topology, and harder graph properties), do LLMs benefit more from compact representations or from schema-rich, human-readable, or diagrammatic representations?
\textbf{RQ1: Representation selection.}
Which graph representations are the most effective across different tasks, graph types, frontier models, and reasoning effort?
\textbf{RQ2: Semantic versus structural representation.}
Are the most effective representations aligned with the task type, with semantically rich representations favoring semantic extraction and connectivity-focused representations favoring structural reasoning?
\textbf{RQ3: Reasoning capacity and cost.} As model reasoning strength increases, does serialization become less important, and how does the preferred representation move along the accuracy--token-cost frontier?
%We study which representation works best for a given task, model, graph type, and token budget.\textbf{with repeated-run variation much smaller than between-format effects}. \textbf{On GraphQA-Hardened, a query-dependent format selector improves accuracy by 9.52 percentage points over the best fixed format and by 11.43 points over task-wise routing, at $3.71\times$ the API cost of the global fixed-best policy}

We make three contributions: 
\textbf{First, we introduce a broad empirical evaluation of graph representations for frontier LLMs.} We compare $15$ representations spanning semantic-web formats, constraint-style renderings, structured natural language, tabular views, diagrammatic formats, compact connectivity encodings, and verbalization of graph structures.
\textbf{Second, we evaluate representation choices across both ontology-driven semantic knowledge extraction and structural graph reasoning.} We use Text2KGBench \citep{mihindukulasooriya2023t2kbench} for ontology-based extraction and extend GraphQA \citep{fatemi2024talklikeagraph} with harder graph sizes, structural operators, and topology variation. 
\textbf{Third, we analyze graph representation selection as a model- and cost-aware orchestration decision.} We evaluate nine frontier LLM configurations, including reasoning-enabled models, and measure both task accuracy and token-cost behavior.  We show that a query-dependent selector model can increase the average performance by $9.52$\,pp over the best fixed format. Together, these contributions position graph serialization as an explicit, dynamic system design choice for KG-augmented LLM pipelines rather than a static prompt-formatting detail.

\section{Related Work}
\label{sec:related_work}
Studies leveraging KGs include \citep{sun2024tog, ma2025tog2}, and constrained KG reasoning frameworks \citep{zhou2025reflection,luo2025gcr}, which have demonstrated improved retrieval, traversal, decoding, and answer generation over graph contexts. Ontology-guided extraction and schema-constrained KG construction methods \citep{mihindukulasooriya2023t2kbench,zhang2024edc} require models to use both semantic labels and topology. Since these systems ultimately prompt LLMs with serialized graphs, they can directly benefit from graph representations that are more efficient.
%including Graph-RAG \citep{edge2024graphrag,peng2024graphrag},
%\paragraph{Graph-to-text interfaces for LLM reasoning.}
Talk-Like-a-Graph (TLAG) \citep{fatemi2024talklikeagraph} showed that graph representations can affect accuracy by 4.8--61.8\,pp on synthetic structural reasoning tasks. GraphArena \citep{tang2025grapharena} broadens graph reasoning evaluation but largely fixes the input representation. NLGraph \citep{wang2023nlgraph} evaluates graph problems stated in natural language; GraphInstruct \citep{luo2025graphinstruct} targets instruction tuning; and GraphToken \citep{perozzi2024graphtoken} bypasses textual representations with learned graph-token embeddings. 
LLM-KG-Bench 3.0 \citep{llmkgbench2025} evaluates LLMs on RDF and SPARQL generation, but gives limited attention to graph representation format. Closest to our setting, KG-LLM-Bench \citep{markowitz2025kgllmbench} shows that textualizing a knowledge graph matters: across KG understanding tasks, language models, and textualization strategies, representation choice produces $17.5$\,pp performance variation. We view this result as the starting point for studying graph-to-text interfaces; however, KG-LLM-Bench does not answer which representational properties matter when the design space includes semantic-web formats, constraint-oriented ontology views, diagrammatic formats, tabular renderings, compact structural encodings, and prose-like representations. Nor does it test whether textualization sensitivity persists for frontier reasoning LLMs, whose stronger instruction following and long-context reasoning could plausibly make representation choice less important. In addition to addressing these questions, our study approaches graph serialization as a model, task, and cost-dependent orchestration decision and proposes a dynamic selector model.

\section{Method}
\label{sec:method}

% Renderers grouped by representational emphasis; in body as a
% two-column-spanning table* so the representation format cells don't wrap.
\begin{table}[t]
\fontsize{7.5pt}{9pt}\selectfont
\centering
\setlength{\tabcolsep}{2pt}
\begin{tabular}{l l}
\toprule
%emphasis & examples \\
%\midrule
formal semantics (RDF/OWL)    & \texttt{turtle}, \texttt{jsonld}, \texttt{shacl\_lite}\\
structured natural language   & \texttt{text2kgbench}, \texttt{entity\_centric} \\
 & \texttt{relation\_indexed} \\
tabular                       & \texttt{markdown\_table}, \texttt{toon} \\
visual / diagrammatic         & \texttt{mermaid}, \texttt{dot} \\
connectivity-focused          & \texttt{adjacency}, \texttt{edge\_list}, \texttt{structured} \\
prose                         & \texttt{nl\_summary}, \texttt{friendship} \\
\bottomrule
\end{tabular}
\caption{Graph formats by semantic explicitness.}
\label{tab:renderers-axis}
\end{table}

\noindent\textbf{Representations.}
We evaluate 15 serialization formats, grouped into six descriptive families (Tab.~\ref{tab:renderers-axis}):
RDF and constraint-oriented Semantic Web formats (\texttt{turtle} \citep{w3c2014turtle}, \texttt{jsonld} \citep{w3c2020jsonld}, and the SHACL-derived \texttt{shacl\_lite} \citep{knublauch2017shacl});
structured natural-language formats (\texttt{text2kgbench} \citep{mihindukulasooriya2023t2kbench}, \texttt{entity\_centric}, and \texttt{relation\_indexed});
tabular formats (\texttt{markdown\_table} and \texttt{toon});
textual diagram languages (\texttt{mermaid} and \texttt{dot});
connectivity-focused encodings (\texttt{adjacency}, \texttt{edge\_list}, and \texttt{structured});
and prose formats (\texttt{nl\_summary} and \texttt{friendship} \citep{fatemi2024talklikeagraph}).
These representation formats differ in how they expose graph information to LLMs and range from semantically explicit representations that expose concept types, relations, and constraints to compact encodings that primarily expose graph connectivity.

\noindent\textbf{Tasks and metrics.}
\emph{Ontology-based knowledge extraction.}
We evaluate the 2,014-sentence test split of Text2KGBench \citep{mihindukulasooriya2023t2kbench}, drawn from the full \texttt{DBPEDIA\_WEBNLG} benchmark spanning 19 ontologies. It pairs each sentence with a serialized ontology and asks the model to extract all supported (subject, relation, object) triples. The task requires semantic understanding of the sentence and ontology-aware reasoning over concept types and relations.

\emph{Structural graph reasoning.}
GraphQA \citep{fatemi2024talklikeagraph} tests whether LLMs can recover graph properties such as edges, degrees, neighbors, counts, paths, and cycles, from textual graph encodings. Such tasks relate to multi-hop KG reasoning, where answers depend on paths and neighborhoods \citep{chen2024llm}. We evaluate three regimes. \textbf{GraphQA} contains 11 tasks: $9$ original structural tasks from TLAG together with $2$ summary tasks we add (\texttt{num\_components}, \texttt{connectivity}), over Erd\H{o}s--R\'enyi graphs with $n\!\in\!\{8,15,25,35,50\}$. \textbf{GraphQA-Hardened} increases graph size to $n\!\in\!\{35,50,75\}$ and adds seven operators that are harder to infer from local edge inspection: $k$-core size \citep{seidman1983kcore}, graph diameter, $k$-shortest path length \citep{yen1971ksp}, local clustering coefficient \citep{watts1998sw}, high-degree-neighbor count, maximum clique size \citep{bron1973bk,tomita2006pivot}, and top betweenness-centrality node \citep{brandes2001betweenness}. \textbf{GraphQA-Topology} fixes the graph size at $n\!=\!50$ and varies the generator across Erd\H{o}s--R\'enyi, Barab\'asi--Albert, stochastic block model, star, complete, and path graphs \citep{erdos1959random,barabasi1999emergence,holland1983sbm}. These extensions let us separate representation effects under larger graphs, harder operators, and different connectivity patterns.

\emph{Reasoning over retrieved Wikidata subgraphs.}
We add an experiment over retrieved subgraphs from Wikidata, using the same models, prompting setup, and 15 graph representations. We select 10 Wikidata seed entities spanning diverse domains, including films, scientists, cities, and companies. For each seed, we retrieve its two-hop neighborhood and capped the resulting context at 450 triples. We generated 110 questions covering relation-specific aggregations, typed entity aggregations, and multi-hop reachability. Examples include: \emph{"How many awards has the scientist Rowan received?", "How many films have at least one director?", "How many distinct entities can be reached from the company Meridian?"}. Entity labels are pseudonymized before serialization, while predicate labels are retained. Each gold answer was computed directly from the canonical subgraph before serialization. Thus, all 15 representations contain the same graph facts and share the same gold answer.

\noindent\textbf{Models and controls.}
We evaluate nine model configurations: \m{gpt-5.4} at the default and \texttt{high} reasoning-effort settings \citep{openai2026gpt54}; \m{gpt-5.4-mini} \citep{openai2026gpt54mini}; \m{o3} \citep{openai2025o3}; \m{claude-sonnet-4-6} with and without extended thinking \citep{anthropic2026sonnet46}; \m{claude-haiku-4-5} \citep{anthropic2025haiku45}; and the reasoning and non-reasoning variants of \m{grok-4-fast} \citep{xai2025grok4fast}. Within each model--task instance, all formats use the same instruction, underlying graph facts, question, seed, node identifiers, node and edge ordering, and task wording; only the serialization changes. Temperature is $0$ for all configurations except thinking-enabled \m{claude-sonnet-4-6}, which uses the API default of $1$. To assess repeated-run stability, we additionally rerun all 15 formats three times with caches disabled on fixed GraphQA-Hardened and GraphQA-Topology evaluation sets, using \m{gpt-5.4} and \m{gpt-5.4-mini}.

\noindent\textbf{Adaptive Format Selection.}
%We test whether instance-specific graph information improves serialization choice beyond global and task-level accuracy priors. 
We design a query-conditioned, instance-specific selector for graph serialization. We show that it improves serialization choice beyond global and task-level accuracy priors.
We construct disjoint, independently seeded calibration and test splits of GraphQA-Hardened (105 questions each) and use \m{gpt-5.4} as both the serialization selector and answer model. Calibration accuracies define per-task, per-format priors and a five-format candidate set. At test time, five parallel fixed-prompt scoring calls (one per candidate) receive the task, format identity, calibration prior, and graph rendered in that format. The highest-scoring rendering is passed to a single answer call; neither selector rationales nor calibration examples are exposed to the answer model. We compare against the best fixed format on the held-out test set, a calibration-only task router, and five-format parallel majority voting, reporting accuracy, latency, and end-to-end inference cost.

\section{Results}
\label{sec:results}

% Headline table --- per-(model, benchmark) best representation, fixed-best score, and per-format spread.
% Cell format: ``\texttt{fmt} score/spread''.  Text2KGBench: Fuzzy F1 @ 0.60 (Hopcroft-Karp).  GraphQA: exact-match accuracy (pp).
\begin{table*}[t]
\centering
\setlength{\tabcolsep}{3pt}
\resizebox{\textwidth}{!}{%
\begin{tabular}{l c c c c}
\toprule
model & Text2KGBench & GraphQA & GraphQA-Hardened & GraphQA-Topology \\
\midrule
\m{gpt-5.4-high-reasoning}     & \texttt{turtle} $54.7$/$1.9$                       &
                                \texttt{structured} $100.0$/$0.6$                   &
                                ---                                                 &
                                --- \\
\m{gpt-5.4}                    & \texttt{mermaid} $52.7$/$3.3$                      &
                                \texttt{shacl\_lite} $93.3$/$4.8$                   &
                                \texttt{nl\_summary} $64.2$/$12.6$                  &
                                \texttt{jsonld} $90.3$/$15.3$ \\
\m{gpt-5.4-mini}               & \texttt{shacl\_lite} $50.6$/$2.6$                  &
                                \texttt{jsonld} $79.4$/$8.5$                        &
                                \texttt{jsonld} $41.0$/$8.4$                        &
                                \texttt{jsonld} $58.3$/$16.7$ \\
\m{o3}                         & \texttt{shacl\_lite} $54.9$/$2.1$                  &
                                \texttt{adjacency} $98.2$/$2.4$                     &
                                \texttt{nl\_summary} $87.4$/$11.6$                  &
                                \texttt{structured} $90.3$/$9.7$ \\
\m{claude-sonnet-4-6-thinking} & \texttt{jsonld} $53.0$/$2.9$                       &
                                 \texttt{text2kgbench} $97.0$/$3.6$                 &
                                 \texttt{text2kgbench} $83.2$/$4.2$                 &
                                 \texttt{jsonld} $95.8$/$5.5$ \\
\m{claude-sonnet-4-6}          & \texttt{jsonld} $52.2$/$2.3$                       &
                                \texttt{nl\_summary} $94.5$/$2.4$                   &
                                \texttt{entity\_centric} $87.4$/$13.7$              &
                                \texttt{relation\_indexed} $84.7$/$6.9$ \\
\m{claude-haiku-4-5}           & \texttt{structured} $50.7$/$3.9$                   &
                                \texttt{shacl\_lite} $90.9$/$7.9$                   &
                                \texttt{adjacency} $72.6$/$28.4$                    &
                                \texttt{nl\_summary} $80.6$/$16.7$ \\
\m{grok-4-fast-reasoning}      & \texttt{shacl\_lite} $48.8$/$1.4$                  &
                                \texttt{nl\_summary} $97.0$/$4.8$                   &
                                \texttt{entity\_centric} $84.2$/$15.8$              &
                                \texttt{adjacency} $95.8$/$16.7$ \\
\m{grok-4-fast-non-reasoning}  & \texttt{mermaid} $49.9$/$1.3$                      &
                                \texttt{adjacency} $81.8$/$17.0$                    &
                                \texttt{shacl\_lite} $44.2$/$21.1$                  &
                                \texttt{nl\_summary} $66.7$/$13.9$ \\
\bottomrule
\end{tabular}}
\caption{Best format per (model, benchmark) as ``score/spread''.  Text2KGBench: F1; GraphQA: exact-match accuracy (pp).  Spread~$=$~max~$-$~min across formats (pp).}
\label{tab:headline}
\end{table*}

We compare graph formats on Text2KGBench and GraphQA and evaluate their effects on task performance, token budgets, and inference costs.
%We organize our results around the research questions introduced in Section~\ref{sec:introduction}. Using six reasoning LLMs, we evaluate which graph representations perform best on Text2KGBench and GraphQA, varying representation richness from formal semantics to connectivity-only encodings. We then analyze token efficiency to determine whether the strongest representations are also practical choices under context-window and inference-cost constraints.

\begin{figure*}[t!]
\centering

\includegraphics[width=0.98\linewidth]{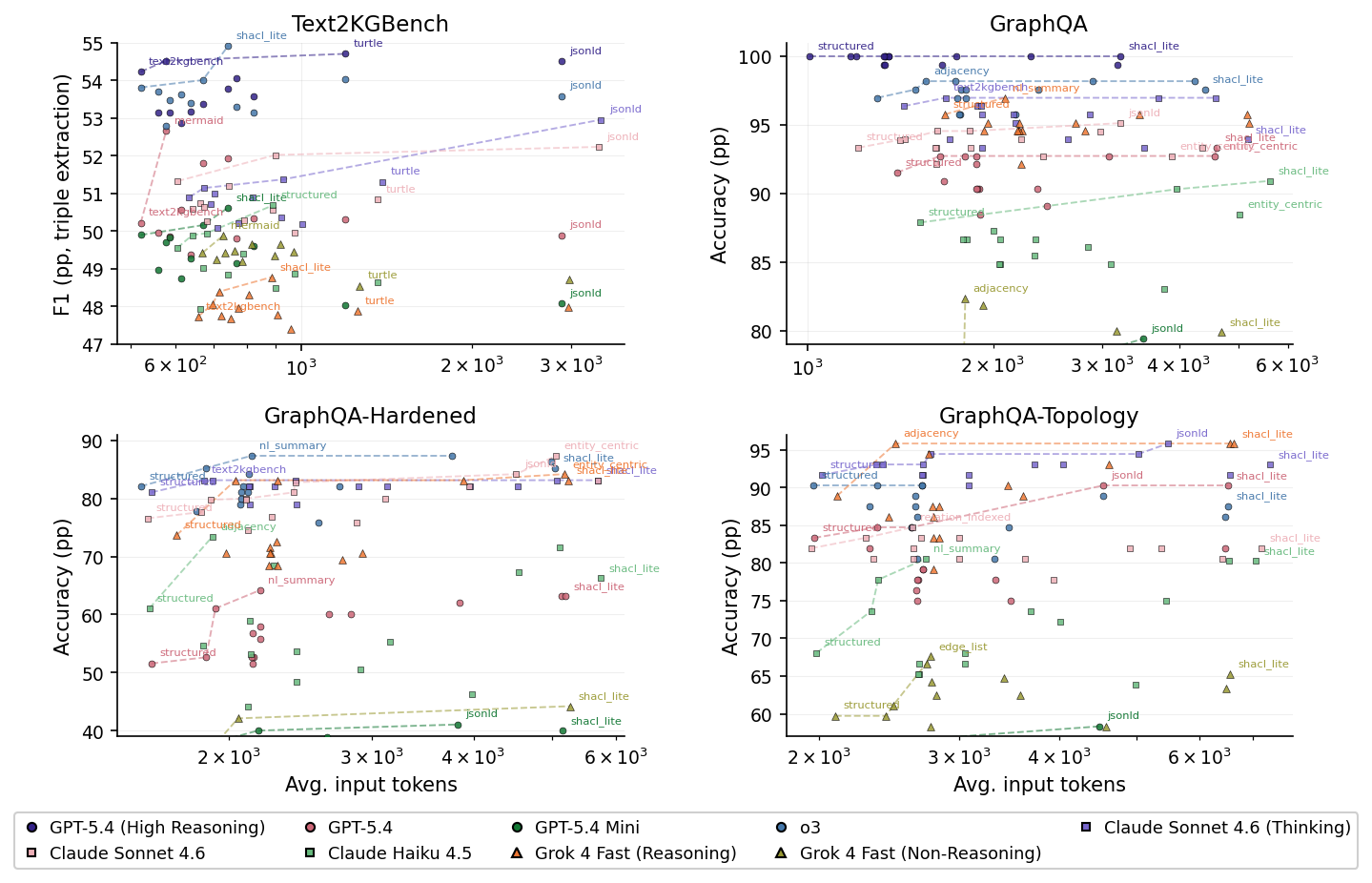}
\caption{Score vs.\ input-token cost across the four
benchmarks. Per-model Pareto fronts dashed.}
\label{fig:01-pareto}
\end{figure*}

\begin{table*}[t]
\centering
\small
\setlength{\tabcolsep}{4.5pt}
\renewcommand{\arraystretch}{1.06}
\begin{tabular*}{\textwidth}{@{\extracolsep{\fill}}l cc rr rrr@{}}
\toprule
& \multicolumn{2}{c}{LLM calls/query}
& \multicolumn{4}{c}{Resources/query}
& \multicolumn{1}{c}{Outcome} \\
\cmidrule(lr){2-3}
\cmidrule(lr){4-7}
\cmidrule(l){8-8}
Policy
  & Selector & Generator
  & Input tok. & Output tok.
  & Rel. cost & Latency
  & Acc. (\%) \\
\midrule
Global fixed-best
  & $0$ & $1$ & $4{,}667$ & $296$ & $1.00\times$ & $5.3$\,s & $66.67$ \\
Task-wise router
  & $0$ & $1$ & $5{,}124$ & $305$ & $1.08\times$ & $5.6$\,s & $64.76$ \\
MV-5 (parallel)
  & $0$ & $5$ & $27{,}231$ & $1{,}573$ & $5.69\times$ & $9.1$\,s & $71.43$ \\
\midrule
\textbf{Adaptive selector}
  & $5$ & $1$ & $17{,}696$ & $1{,}034$ & $3.71\times$ & $9.4$\,s
  & $\mathbf{76.19}$ \\
\bottomrule
\end{tabular*}
\caption{End-to-end policy comparison on the same 105 held-out
GraphQA-Hardened questions. Counts cover the complete pipeline, and API
cost is normalized to global fixed-best. The five selector scores and the
five majority voting (MV-5) generations are issued in parallel when measuring latency.}
\label{tab:selector-policy}
\end{table*}

\noindent\textbf{Ontology-based knowledge extraction.}
\label{sec:results:c1}
%Text2KGBench pairs a free-text sentence with a serialized ontology and asks each LLM to output all \texttt{(subject, relation, object)} triples, extracting the knowledge graph for the ontology. 
In this experiment we serialize the ontology into each candidate format; \texttt{text2kgbench} denotes the original compact Text2KGBench serialization. All nine configurations obtain $45$--$55$\,F1 (Tab.~\ref{tab:headline}, leftmost column; Fig.~\ref{fig:01-pareto}, top-left). 
Our best extraction result, \m{o3} with \texttt{shacl\_lite} ($54.9$\,F1), is $\sim\!25$\,pp above the \citet{mihindukulasooriya2023t2kbench} baselines. The variation in F1 score due to the graph representation ranges between $1.3$--$3.9$ percentage points. Most configurations perform best using semantically explicit representations. This result aligns with expectations, as semantically rich representations can better support tasks requiring semantic understanding of concept types and relations. On Text2KGBench's $19$ WebNLG domain ontologies, frontier models leave modest room for improvements; gains stem from model capacity rather than serialization choice.\\
\noindent\textbf{Structural graph reasoning.}
\label{sec:results:c2}
%We re-implement the Talk-Like-a-Graph GraphQA protocol \citep{fatemi2024talklikeagraph} as our base structural reasoning setting, using freshly generated Erd\H{o}s--R\'enyi graphs. Our GraphQA setting keeps nine Talk-Like-a-Graph tasks, adds two structural-summary tasks (\texttt{num\_components} and \texttt{connectivity}), and asks questions about edges, degrees, connectivity, shortest paths, and cycles. To avoid saturation on frontier models, we introduce two harder variants: GraphQA-Hardened increases graph size ($n\!\in\!\{35,50,75\}$) and adds more demanding structural queries, while GraphQA-Topology varies the graph generator across six families (Appendix~\ref{appx:spread}, Tab.~\ref{tab:topo-params}).
For structural graph reasoning (Tab.~\ref{tab:headline}) we find that  \m{gpt-5.4-high} solves GraphQA with 100\% accuracy using the compact \texttt{structured} format, and it is less sensitive to format representation. Format sensitivity increases as LLM reasoning level drops, rising from $4.8$\,pp on default-tier \m{gpt-5.4} to $17.0$\,pp on \m{grok-4-fast-non-reasoning}. Harder settings further widen the gap for all models: $4.2$--$28.4$\,pp on GraphQA-Hardened and $5.5$--$16.7$\,pp on GraphQA-Topology. 
%Graph topology also matters: Barab\'asi--Albert graphs are $+28$\,pp easier than Erd\H{o}s--R\'enyi graphs at matched edge count on the hardest tasks (Tab.~\ref{tab:topology-means}), suggesting that ER-only evaluations understate variation on heavy-tailed real-world graphs \citep{barabasi1999emergence,clauset2009powerlaw}.
Results in (Tab. \ref{tab:headline}) show that no single format dominates structural reasoning. The best graph representation is model- and task-dependent.
Representation preferences can change with reasoning tier even within a single model family. For the GraphQA task \m{gpt-5.4-mini} favors \texttt{jsonld} ($79.4$\,pp, $3{,}501$ tokens), \m{gpt-5.4} favors \texttt{shacl\_lite} ($93.3$\,pp, $4{,}608$ tokens), and \m{gpt-5.4-high} reaches $100.0$\,pp with compact \texttt{structured} ($1{,}008$ tokens). We use a fixed prompt for controlled comparisons; it is not intended to be the optimal prompt for every model family.\\
\noindent\textbf{Retrieved Wikidata subgraphs.}
\label{sec:results:wikidata}
In a targeted evaluation using \m{gpt-5.4} and \m{gpt-5.4-mini}, we evaluate 110 questions over 10 Wikidata-derived subgraphs; both configurations rank \texttt{jsonld} first and \texttt{turtle} second. \m{gpt-5.4} scores $41.8\%$ and $40.0\%$, compared with $13.6\%$ for \texttt{adjacency}, a $28.2$ percentage-point spread. \m{gpt-5.4-mini} scores $30.9\%$ and $29.1\%$, compared with $6.4\%$ for \texttt{nl\_summary}, a $24.5$ percentage-point spread. The similarly large spreads for \m{gpt-5.4} and \m{gpt-5.4-mini} show that representation sensitivity persists beyond synthetic GraphQA and across model capability levels on real-world Wikidata subgraphs.\\
\noindent\textbf{Representation characteristics and token efficiency.}
\label{sec:results:pareto}
Fig.~\ref{fig:01-pareto} plots scores against average input tokens; dashed lines show the accuracy vs. token cost tradeoff (per-model Pareto fronts). On Text2KGBench (top-left), scores fall within a narrow F1 band. \texttt{shacl\_lite}, \texttt{text2kgbench} and \texttt{mermaid} are token-efficient while achieving high accuracy. In contrast, \texttt{turtle} and \texttt{jsonld} have a substantially higher token cost. On GraphQA (top-right), reasoning-enabled configurations generally occupy the high-accuracy region across a wide token range, and compact formats such as \texttt{adjacency} and \texttt{structured} achieve high accuracy at low-cost, even for harder tasks. On harder tasks, GraphQA-Hardened and GraphQA-Topology (bottom row), format choice affects both accuracy and token cost. For LLMs with default or reduced reasoning capabilities, verbose representations may be justified to increase accuracy.\\
\noindent\textbf{Adaptive Format Selection.}
\label{sec:results:adaptive-selection}
The task-wise router using disjoint calibration priors reaches 64.76\%, compared with 66.67\% for global fixed-best, 71.43\% for five-format majority voting, and 76.19\% for the fully adaptive selector (Tab.~\ref{tab:selector-policy}). The selector therefore exceeds task-wise routing by 11.43 points, showing that task-level priors alone do not explain its gain, and exceeds majority voting by 4.76 points. Relative to global fixed-best, the selector gains $\mathbf{+9.52}$ percentage points, with a paired-BCa-bootstrap 95\% CI of $[+1.90,+16.19]$ and a mid-$p$ McNemar test yielding $p=0.013$. The post-hoc oracle ceiling over the same five-format candidate slate is 82.86\%.

The complete selector pipeline makes five parallel selector calls and one generator call per query, averaging $17{,}696/1{,}034$ input/output tokens. It incurs $3.71\times$ the API cost of global fixed-best and has an end-to-end latency of 9.4\,s. Parallel MV-5 has similar latency at 9.1\,s but lower accuracy and higher cost, reaching $71.43\%$ at $5.69\times$ the fixed-best cost. These results demonstrate the feasibility of query-dependent selection on unseen, separately seeded graphs drawn from the same GraphQA-Hardened distribution. Broader transfer across datasets, graph distributions, and selector--generator model families remains future work.

\section{Conclusion}
\label{sec:conclusion}

We systematically evaluate 15 graph serializations across nine frontier LLM configurations with varied reasoning effort. Serialization effects are modest for ontology-guided extraction (per-format F1 spread up to $3.9\,\mathrm{pp}$) but substantially larger for structural reasoning: up to 28.4 points on controlled GraphQA regimes, and 24.5--28.2 points on retrieved Wikidata contexts.

Graph serialization is therefore a measurable systems decision, not a cosmetic prompt choice. Across the evaluated settings, formats including \texttt{shacl\_lite}, \texttt{mermaid}, and \texttt{jsonld} can be strong candidates, but no representation is a universal default; the accuracy--token-cost tradeoff depends on the model, task, and graph regime. Our in-distribution serialization-selector study achieves $9.52$\,pp improvement over the best fixed format on GraphQA Hardened. More broadly, we suggest that KG-grounded LLM systems should report, evaluate, and, where worthwhile, adapt their graph-to-text interface as a dynamic, model- and task-aware orchestration choice.
\\

% --------- Limitations, Ethics, Acks: do NOT count toward 5-page limit.
\section*{Limitations}

This study focuses on graph and ontology serialization as text for LLMs; its findings may not generalize to graph-native methods, learned graph tokens, tools, constrained decoding, or end-to-end Graph-RAG systems. Evaluation is limited to Text2KGBench, synthetic GraphQA, and a two-model Wikidata study over fixed retrieved subgraphs. We do not assess retrieval quality, larger contexts, temporal or probabilistic graphs, retrieval errors, or open-ended KGQA.

Our 15 hand-designed serializers are not exhaustive. They vary in explicitness, grouping, ordering, verbosity, and token cost. Paired comparisons control for model, graph facts, question, and prompt. Finally, all evaluated models are closed-weight commercial systems, so we cannot determine the effect of prior exposure to different serialization formats.

\section*{Ethical Considerations}
\label{sec:ethics}

The experiments use the publicly released WebNLG split of Text2KGBench \citep{mihindukulasooriya2023t2kbench}, synthetic graphs generated for our GraphQA extensions, and cached subgraphs retrieved from Wikidata \citep{vrandecic2014wikidata}. We collect no new human-subject data and submit no private, confidential, or organizational data to model endpoints.

Model requests are sent through access-controlled enterprise endpoints. The submitted content consists only of benchmark sentences, serialized ontologies or graphs, and task instructions.
\section*{Acknowledgments}
\label{sec:acknowledgments}

The authors used generative AI tools for limited language editing, rewriting suggestions, and assistance in improving exposition. All scientific content, experimental design, code, analyses, claims, citations, and final text were reviewed, verified, and approved by the human authors.

% --------- References (unlimited).
\bibliography{custom}

% --------- Appendix (after references; reviewers not required to read).
\newpage
\appendix
% =====================================================================
% Appendix.  Appears after \appendix, after references.  Per ARR/EMNLP:
% appendices come after the references in the submitted PDF, do not
% count toward the page limit, and reviewers are not required to
% consider them.
%
% Order: A) GraphQA detail (task inventory, topology generators,
%           per-slot ceiling, per-format spread, spread-vs-headroom);
%        B) Worked format-selector run (gpt-5.4 by gpt-5.4 on
%           GraphQA-Hardened);
%        C) Text2KGBench detail;
%        D) Statistical pipeline;
%        E) Representation format inventory and prompts.
% =====================================================================
% A  GraphQA detail (task inventory, topology generators, ceiling,
%                    per-format spread, spread-vs-headroom figure)
% =====================================================================
\section{GraphQA}
\label{appx:spread}

In this section we provide additional information on the GraphQA tasks, focusing on the extensions we implemented to make this task harder.

%This appendix backs the structural-reasoning claims of \S\ref{sec:results:c2}: the task inventory (Tab.~\ref{tab:graphqa-tasks}), the topology generators (Tab.~\ref{tab:topo-params}), the per-slot ceiling (Tab.~\ref{tab:ceiling}), the per-format spread (Tab.~\ref{tab:spread}), and the spread-vs-headroom figure (Fig.~\ref{fig:03-spread}).

\paragraph{Task inventory and the three regimes.}
The three regimes differ along \emph{task set}, \emph{graph size}, and \emph{graph generator} (Tab.~\ref{tab:graphqa-tasks}).

\textbf{GraphQA} reproduces the $9$ Fatemi--Perozzi tasks \citep{fatemi2024talklikeagraph} and adds two structural summaries (\texttt{num\_components}, \texttt{connectivity}) at $n\!\in\!\{8,15,25,35,50\}$ under $p\!\sim\!U[0,1]$ Erd\H{o}s--R\'enyi density.

\textbf{GraphQA-Hardened} layers $7$ harder operators on top at $n\!\in\!\{35,50,75\}$ under a calibrated $p\!=\!\min(0.5,\,2.5\log_{10} n/n)$ so $k$-core, diameter, and max-clique stay non-trivial.  The added operators cover: $k$-core peeling \citep{seidman1983kcore}, diameter via all-pairs reachability \citep{watts1998sw}, $k$-shortest paths via Yen \citep{yen1971ksp}, max-clique via Bron--Kerbosch / Tomita pivot \citep{bron1973bk,tomita2006pivot}, top-betweenness \citep{brandes2001betweenness}, local clustering, and high-degree-neighbour count.

\textbf{GraphQA-Topology} fixes $n\!=\!50$ at the calibrated density, restricts to the $4$ bounded-answer operators (\texttt{kth\_shortest\_path}, \texttt{local\_clustering}, \texttt{neighbour\_degree\_ge}, \texttt{betweenness\_top1}), and varies the generator over six families (Tab.~\ref{tab:topo-params}).  Reference examples are computed on the $5$-node graph from App.~\ref{appx:renderers}.

\begin{table*}[h!]
\small
\centering
\setlength{\tabcolsep}{4pt}
\begin{tabular}{l p{0.42\linewidth} l l}
\toprule
task & question template & ref.\ example & in \\
\midrule
\multicolumn{4}{l}{\textit{GraphQA (11 tasks; $9$ adapted from \citealp{fatemi2024talklikeagraph} $+\,2$ added).}} \\[2pt]
\texttt{node\_count}        & ``How many nodes are in the graph?''                                                                    & $5$              & G \\
\texttt{edge\_count}        & ``How many edges are in the graph?''                                                                    & $5$              & G \\
\texttt{node\_degree}       & ``What is the degree of node $v$?''                                                                     & $\deg(3)\!=\!3$  & G \\
\texttt{edge\_existence}    & ``Is there an edge directly between $a$ and $b$? Yes/No.''                                              & $\{0,4\}\!\to$ No & G \\
\texttt{connectivity}       & ``Is there a path from $a$ to $b$? Yes/No.''                                                            & $0\!\to\!4$ Yes  & G \\
\texttt{connected\_nodes}   & ``List all direct neighbours of $v$, comma-separated.''                                                 & $N(3)\!=\!1,2,4$ & G \\
\texttt{cycle\_check}       & ``Does the graph contain at least one cycle? Yes/No.''                                                  & Yes              & G \\
\texttt{shortest\_path}     & ``Length of shortest $a\!\to\!b$ path? $-1$ if disconnected.''                                          & $d(0,4)\!=\!3$   & G \\
\texttt{triangle\_counting} & ``How many unordered triangles?''                                                                       & $0$              & G \\
\texttt{num\_components}    & ``How many connected components?'' \emph{(added)}                                                       & $1$              & G \\
\texttt{disconnected\_nodes}& ``List every node \emph{not} reachable from $v$, comma-separated.''                                     & none             & G \\
\midrule
\multicolumn{4}{l}{\textit{GraphQA-Hardened (7 tasks; added on top of the GraphQA $11$).}} \\[2pt]
\texttt{kcore\_size}        & ``How many nodes belong to the $2$-core?''                                                              & $4$              & H \\
\texttt{graph\_diameter}    & ``Within the largest component, what is the diameter?''                                                 & $3$              & H \\
\texttt{max\_clique\_size}  & ``Size of the largest clique?''                                                                         & $2$              & H \\
\texttt{kth\_shortest\_path}& ``Length of the second-shortest simple $a\!\to\!b$ path? $-1$ if none.''                                & $0\!\to\!4\!:\!-1$ & H,T \\
\texttt{betweenness\_top1}  & ``Which node has the HIGHEST betweenness centrality? Ties: smallest id.''                               & $3$              & H,T \\
\texttt{local\_clustering}  & ``Local clustering coefficient of $v$? (2 d.p.)''                                                       & $C(3)\!=\!0.00$  & H,T \\
\texttt{neighbour\_degree\_ge} & ``How many neighbours of $v$ have degree $\!\ge\!k$, where $k\!=\!\lceil\bar d\rceil$?''             & $v\!=\!3,k\!=\!2\!:\!2$ & H,T \\
\bottomrule
\end{tabular}
\caption{The $18$ task templates ($11$ GraphQA + $7$ Hardened).  ``in'': \textbf{G}~=~GraphQA, \textbf{H}~=~Hardened, \textbf{T}~=~Topology.  Topology keeps only the $4$ Hardened operators with a bounded integer / 2-dp answer; reference examples use the $5$-node graph of App.~\ref{appx:renderers}.}
\label{tab:graphqa-tasks}
\end{table*}

\paragraph{Topology generators.}
The six generators at $n\!=\!50$ span $49$--$1{,}225$ edges (Tab.~\ref{tab:topo-params}).  Erd\H{o}s--R\'enyi (ER) and Barab\'asi--Albert (BA) graphs differ by $\sim\!15\%$ in mean edge count, so the BA-vs-ER contrast is at approximately matched density; stochastic block model (SBM) and complete confound topology with edge count, and star/path are degenerate by construction.

\begin{table*}[!t]
\small
\centering
\setlength{\tabcolsep}{8pt}
\begin{tabular}{l l l l}
\toprule
generator   & parameters & degree distribution & edge count at $n\!=\!50$ \\
\midrule
ER $G(n,p)$ & $p\!=\!\min(0.5,\,2.5 \log_{10} n / n)\!\approx\!0.085$ & binomial, sharp peak                & $\sim\!85$ \\
BA          & $m\!=\!2$ edges per new node                            & power-law $P(k)\!\sim\!k^{-3}$       & $\approx\!98$ \\
SBM         & $2$ communities, $p_{\text{in}}\!=\!0.55$, $p_{\text{out}}\!=\!0.03$ & bimodal                  & $\sim\!360$ \\
star        & $1$ hub, $n\!-\!1$ leaves                               & $\{1,\, n\!-\!1\}$                   & $49$ \\
complete    & all-to-all                                              & $\{n\!-\!1\}$                        & $1{,}225$ \\
path        & $0\!-\!1\!-\!\ldots\!-\!(n\!-\!1)$                      & $\{1,\, 2\}$                         & $49$ \\
\bottomrule
\end{tabular}
\caption{GraphQA-Topology generators at $|V|\!=\!50$.  Density-matched comparisons across all six are not meaningful; the load-bearing contrast (\S\ref{sec:results:c2}) is BA vs.\ ER at $\sim\!15\%$ density mismatch.}
\label{tab:topo-params}
\end{table*}

\paragraph{Full per-(model, format) accuracy.}
Tabs.~\ref{tab:gqa-full}--\ref{tab:gqa-topology-full} list the raw accuracy of every (model, format) cell on the three GraphQA regimes.  Rows are models (full canonical names); columns are representation formats (header rotated $60^{\circ}$).  Bold marks the per-row (per-model) maximum.

\begin{table*}[!t]
\fontsize{7}{8.4}\selectfont
\centering
\setlength{\tabcolsep}{2pt}
\resizebox{\textwidth}{!}{%
\begin{tabular}{l r r r r r r r r r r r r r r r}
\toprule
model & \rotatebox{60}{\texttt{adjacency}} & \rotatebox{60}{\texttt{dot}} & \rotatebox{60}{\texttt{edge\_list}} & \rotatebox{60}{\texttt{entity\_centric}} & \rotatebox{60}{\texttt{friendship}} & \rotatebox{60}{\texttt{jsonld}} & \rotatebox{60}{\texttt{markdown\_table}} & \rotatebox{60}{\texttt{mermaid}} & \rotatebox{60}{\texttt{nl\_summary}} & \rotatebox{60}{\texttt{relation\_indexed}} & \rotatebox{60}{\texttt{shacl\_lite}} & \rotatebox{60}{\texttt{structured}} & \rotatebox{60}{\texttt{text2kgbench}} & \rotatebox{60}{\texttt{toon}} & \rotatebox{60}{\texttt{turtle}} \\
\midrule
\m{gpt-5.4-high-reasoning} & \textbf{100.0} & \textbf{100.0} & \textbf{100.0} & 99.4 & 99.4 & \textbf{100.0} & \textbf{100.0} & \textbf{100.0} & 99.4 & \textbf{100.0} & \textbf{100.0} & \textbf{100.0} & \textbf{100.0} & 99.4 & \textbf{100.0} \\
\m{gpt-5.4} & 90.9 & 88.5 & 92.1 & 92.7 & 90.3 & 92.7 & 90.3 & 90.3 & 92.7 & 90.3 & \textbf{93.3} & 91.5 & 92.7 & 92.7 & 89.1 \\
\m{gpt-5.4-mini} & 75.1 & 72.1 & 73.3 & 77.6 & 71.5 & \textbf{79.4} & 75.1 & 73.9 & 76.4 & 75.8 & 76.4 & 75.1 & 72.7 & 71.5 & 70.9 \\
\m{o3} & \textbf{98.2} & 97.0 & 95.8 & \textbf{98.2} & 95.8 & \textbf{98.2} & 97.0 & 97.6 & \textbf{98.2} & 97.6 & 97.6 & 97.0 & 97.6 & 95.8 & 97.6 \\
\m{claude-sonnet-4-6-thinking} & 93.9 & 95.2 & 95.8 & \textbf{97.0} & 95.8 & \textbf{97.0} & 93.9 & 95.8 & 96.4 & 96.4 & 93.9 & 96.4 & \textbf{97.0} & 93.3 & 93.3 \\
\m{claude-sonnet-4-6} & 93.9 & \textbf{94.5} & 92.1 & 92.7 & 92.7 & \textbf{94.5} & 93.9 & 92.7 & \textbf{94.5} & 93.3 & 93.3 & 93.3 & 93.3 & 93.3 & 93.9 \\
\m{claude-haiku-4-5} & 86.7 & 86.7 & 86.7 & 88.5 & 84.9 & 90.3 & 86.1 & 85.5 & 87.3 & 84.9 & \textbf{90.9} & 87.9 & 86.7 & 84.9 & 83.0 \\
\m{grok-4-fast-reasoning} & 95.2 & 92.1 & 94.5 & 95.8 & 95.2 & 95.8 & 94.5 & 94.5 & \textbf{97.0} & 94.5 & 95.2 & 95.8 & 94.5 & 95.2 & 94.5 \\
\m{grok-4-fast-non-reasoning} & \textbf{81.8} & 66.7 & 66.7 & 77.6 & 67.9 & 80.0 & 69.1 & 69.1 & \textbf{81.8} & 67.9 & 79.4 & 66.7 & 66.7 & 64.8 & 65.5 \\
\bottomrule
\end{tabular}
}
\caption{GraphQA per-format accuracy (\%; ER, $|V|\!\le\!50$).  Bold: per-model maximum.  $n\!=\!165$ questions per (model, format).}
\label{tab:gqa-full}
\end{table*}

\begin{table*}[!t]
\fontsize{7}{8.4}\selectfont
\centering
\setlength{\tabcolsep}{2pt}
\resizebox{\textwidth}{!}{%
\begin{tabular}{l r r r r r r r r r r r r r r r}
\toprule
model & \rotatebox{60}{\texttt{adjacency}} & \rotatebox{60}{\texttt{dot}} & \rotatebox{60}{\texttt{edge\_list}} & \rotatebox{60}{\texttt{entity\_centric}} & \rotatebox{60}{\texttt{friendship}} & \rotatebox{60}{\texttt{jsonld}} & \rotatebox{60}{\texttt{markdown\_table}} & \rotatebox{60}{\texttt{mermaid}} & \rotatebox{60}{\texttt{nl\_summary}} & \rotatebox{60}{\texttt{relation\_indexed}} & \rotatebox{60}{\texttt{shacl\_lite}} & \rotatebox{60}{\texttt{structured}} & \rotatebox{60}{\texttt{text2kgbench}} & \rotatebox{60}{\texttt{toon}} & \rotatebox{60}{\texttt{turtle}} \\
\midrule
\m{gpt-5.4} & 61.1 & 55.8 & 56.8 & 63.2 & 60.0 & 62.1 & 52.6 & 57.9 & \textbf{64.2} & 52.6 & 63.2 & 51.6 & 52.6 & 51.6 & 60.0 \\
\m{gpt-5.4-mini} & 37.9 & 35.8 & 32.6 & 35.8 & 39.0 & \textbf{41.0} & 34.7 & 34.7 & 40.0 & 33.7 & 40.0 & 34.7 & 33.7 & 33.7 & 34.7 \\
\m{o3} & 85.3 & 81.0 & 79.0 & 86.3 & 75.8 & \textbf{87.4} & 82.1 & 84.2 & \textbf{87.4} & 80.0 & 85.3 & 82.1 & 77.9 & 81.0 & 82.1 \\
\m{claude-sonnet-4-6-thinking} & \textbf{83.2} & 79.0 & 79.0 & \textbf{83.2} & 82.1 & 82.1 & 82.1 & \textbf{83.2} & 82.1 & 82.1 & \textbf{83.2} & 81.0 & \textbf{83.2} & 82.1 & 82.1 \\
\m{claude-sonnet-4-6} & 79.0 & 81.0 & 80.0 & \textbf{87.4} & 80.0 & 84.2 & 75.8 & 81.0 & 76.8 & 79.0 & 83.2 & 75.8 & 76.8 & 73.7 & 82.1 \\
\m{claude-haiku-4-5} & \textbf{72.6} & 53.7 & 52.6 & 71.6 & 54.7 & 67.4 & 50.5 & 48.4 & 68.4 & 44.2 & 66.3 & 61.1 & 54.7 & 59.0 & 46.3 \\
\m{grok-4-fast-reasoning} & 83.2 & 68.4 & 71.6 & \textbf{84.2} & 69.5 & 83.2 & 70.5 & 72.6 & 83.2 & 68.4 & 83.2 & 73.7 & 70.5 & 70.5 & 70.5 \\
\m{grok-4-fast-non-reasoning} & 42.1 & 28.4 & 27.4 & 35.8 & 23.2 & 35.8 & 26.3 & 24.2 & 34.7 & 26.3 & \textbf{44.2} & 30.5 & 23.2 & 26.3 & 28.4 \\
\bottomrule
\end{tabular}
}
\caption{GraphQA-Hardened per-format accuracy (\%; $|V|\!\in\!\{35,50,75\}$).  Bold: per-model maximum.  $n\!=\!95$ questions per (model, format).  \m{gpt-5.4-high-reasoning} excluded from this full-format comparison.}
\label{tab:gqa-hardened-full}
\end{table*}

\begin{table*}[!t]
\fontsize{7}{8.4}\selectfont
\centering
\setlength{\tabcolsep}{2pt}
\resizebox{\textwidth}{!}{%
\begin{tabular}{l r r r r r r r r r r r r r r r}
\toprule
model & \rotatebox{60}{\texttt{adjacency}} & \rotatebox{60}{\texttt{dot}} & \rotatebox{60}{\texttt{edge\_list}} & \rotatebox{60}{\texttt{entity\_centric}} & \rotatebox{60}{\texttt{friendship}} & \rotatebox{60}{\texttt{jsonld}} & \rotatebox{60}{\texttt{markdown\_table}} & \rotatebox{60}{\texttt{mermaid}} & \rotatebox{60}{\texttt{nl\_summary}} & \rotatebox{60}{\texttt{relation\_indexed}} & \rotatebox{60}{\texttt{shacl\_lite}} & \rotatebox{60}{\texttt{structured}} & \rotatebox{60}{\texttt{text2kgbench}} & \rotatebox{60}{\texttt{toon}} & \rotatebox{60}{\texttt{turtle}} \\
\midrule
\m{gpt-5.4} & 84.7 & 79.2 & 75.0 & 81.9 & 77.8 & \textbf{90.3} & 77.8 & 79.2 & 84.7 & 76.4 & \textbf{90.3} & 83.3 & 81.9 & 77.8 & 75.0 \\
\m{gpt-5.4-mini} & 48.6 & 50.0 & 52.8 & 48.6 & 50.0 & \textbf{58.3} & 54.2 & 54.2 & 47.2 & 52.8 & 41.7 & 55.6 & 52.8 & 48.6 & 50.0 \\
\m{o3} & \textbf{90.3} & \textbf{90.3} & 80.6 & 86.1 & 80.6 & 88.9 & 86.1 & \textbf{90.3} & 84.7 & 87.5 & 87.5 & \textbf{90.3} & 87.5 & 88.9 & 84.7 \\
\m{claude-sonnet-4-6-thinking} & 93.1 & 90.3 & 93.1 & 91.7 & 93.1 & \textbf{95.8} & 93.1 & 91.7 & 94.4 & 91.7 & 93.1 & 91.7 & 93.1 & 91.7 & 94.4 \\
\m{claude-sonnet-4-6} & 80.6 & 83.3 & 80.6 & 80.6 & 77.8 & 81.9 & 80.6 & 80.6 & 83.3 & \textbf{84.7} & 81.9 & 81.9 & 83.3 & 81.9 & 81.9 \\
\m{claude-haiku-4-5} & 77.8 & 66.7 & 66.7 & 79.2 & 72.2 & 75.0 & 73.6 & 68.1 & \textbf{80.6} & 65.3 & 79.2 & 68.1 & 73.6 & 65.3 & 63.9 \\
\m{grok-4-fast-reasoning} & \textbf{95.8} & 83.3 & 86.1 & \textbf{95.8} & 90.3 & 93.1 & 83.3 & 87.5 & 94.4 & 87.5 & \textbf{95.8} & 88.9 & 86.1 & 79.2 & 88.9 \\
\m{grok-4-fast-non-reasoning} & 61.1 & 55.6 & \textbf{66.7} & 62.5 & 63.9 & 58.3 & 52.8 & 62.5 & \textbf{66.7} & 62.5 & 65.3 & 59.7 & 59.7 & 58.3 & 62.5 \\
\bottomrule
\end{tabular}
}
\caption{GraphQA-Topology per-format accuracy (\%; $|V|\!=\!50$, six generators).  Bold: per-model maximum.  $n\!=\!72$ questions per (model, format).  \m{gpt-5.4-high-reasoning} excluded from this full-format comparison.}
\label{tab:gqa-topology-full}
\end{table*}

% =====================================================================
% B  Retrieved Wikidata validation.
% =====================================================================
\section{Reasoning over Retrieved Wikidata Subgraphs}
\label{appx:wikikg}

To assess whether representation sensitivity extends beyond synthetic GraphQA instances, we construct a controlled evaluation using Wikidata \citep{vrandecic2014wikidata}. We select 10 seed entities from heterogeneous domains, including films, scientists, cities, and companies. For each seed, we retrieve its directed two-hop neighborhood and retain at most 450 triples; the resulting subgraph serves as the canonical evaluation context. We replace entity labels with consistent anonymous identifiers while preserving predicate labels. This retains real-KG relational and topological structure while limiting reliance on memorized entity-specific knowledge.

The evaluation comprises 110 questions: 31 relation-specific aggregation questions, 31 type-aggregation questions, and 48 directed-reachability questions. As a targeted extension to the primary evaluation, we test all 15 serializers with two model configurations, \m{gpt-5.4} and \m{gpt-5.4-mini}. Gold answers are computed from the canonical subgraphs before rendering, and every serializer receives the same underlying subgraph for each question.

\begin{table}[t]
\small
\centering
\setlength{\tabcolsep}{3pt}
\resizebox{\columnwidth}{!}{%
\begin{tabular}{l l r l r l r r}
\toprule
model & best & acc. & second & acc. & worst & acc. & spread \\
\midrule
\m{gpt-5.4}
  & \texttt{jsonld} & 41.8
  & \texttt{turtle} & 40.0
  & \texttt{adj.} & 13.6 & \textbf{28.2} \\
\m{gpt-5.4-mini}
  & \texttt{jsonld} & 30.9
  & \texttt{turtle} & 29.1
  & \texttt{nl\_sum.} & 6.4 & \textbf{24.5} \\
\bottomrule
\end{tabular}}
\caption{Tolerance accuracy (\%) on 110 questions over retrieved Wikidata subgraphs. Spread is the within-model maximum minus minimum across all 15 formats.}
\label{tab:wikikg}
\end{table}

Table~\ref{tab:wikikg} reports the results. Both configurations rank
\texttt{jsonld} first and \texttt{turtle} second. Across all 15 formats, the best-to-worst spread is 28.2 percentage points for \m{gpt-5.4} and 24.5 percentage points for \m{gpt-5.4-mini}. These results demonstrate representation sensitivity in this targeted two-model evaluation on heterogeneous Wikidata-derived subgraphs. Because each canonical subgraph is fixed before serialization, the experiment evaluates reasoning over retrieved graph contexts rather than retrieval quality or an end-to-end Graph-RAG pipeline.

% =====================================================================
% C  Robustness and representation properties.
% =====================================================================
\section{Robustness and Representation Properties}
\label{appx:robustness}

\paragraph{Repeated-run stability.}
We repeated all 15 formats for three independent decodes on fixed held-out sets of 105 GraphQA-Hardened and 120 GraphQA-Topology questions, using \m{gpt-5.4} and \m{gpt-5.4-mini}. All runs used temperature $0$, and response caching was disabled so that repeated prompts were submitted independently. This gives $15 \times 3 \times (105+120) \times 2 = 20{,}250$ model calls. Across the four model--benchmark settings, the mean per-format run-to-run accuracy standard deviation is 1.5 percentage points. The between-format accuracy spread is $7.7\times$ larger on average. Serving-side nondeterminism is therefore substantially smaller than the observed representation effect.

\paragraph{Representation properties.}
The family labels in Tab.~\ref{tab:renderers-axis} are descriptive rather than causal. No individual characteristic determines performance across settings. Rankings vary by task, graph regime, and model configuration; a format that clearly exposes local neighborhoods, for example, need not perform best on a global graph operator. The targeted Wikidata evaluation provides a complementary real-KG case, with both evaluated models favoring the schema-explicit \texttt{jsonld} and \texttt{turtle} formats. We therefore interpret these regularities as associations between information organization and task demands, not as a causal decomposition of representation effects.

% =====================================================================
% D  Text2KGBench detail (per-format bars, per-domain heatmap,
%                         per-model best/worst, prior-art comparison,
%                         raw per-(model, format) F1 grid).
% =====================================================================
\section{Text2KGBench}
\label{appx:t2k}

In this section we show the comparison to the original baseline and detailed scores for every task, model, and representation.

\begin{table*}[h!]
\small
\centering
\setlength{\tabcolsep}{8pt}
\begin{tabular}{l c r r}
\toprule
model (year)                  & best fmt              & F1                & $\Delta$ vs.\ Vicuna 2023 \\
\midrule
Vicuna-13B, 2023              & \texttt{text2kgbench} & $30.0$            & --- \\
Alpaca-LoRA-13B, 2023         & \texttt{text2kgbench} & $25.0$            & $-5.0$ \\
\midrule
\m{gpt-5.4-high-reasoning}     & \texttt{turtle}       & $54.7$            & $+24.7$ \\
\m{gpt-5.4}                    & \texttt{mermaid}      & $52.7$            & $+22.7$ \\
\m{gpt-5.4-mini}               & \texttt{shacl\_lite}  & $50.6$            & $+20.6$ \\
\m{o3}                         & \texttt{shacl\_lite}  & $\mathbf{54.9}$   & $+24.9$ \\
\m{claude-sonnet-4-6-thinking} & \texttt{jsonld}       & $53.0$            & $+23.0$ \\
\m{claude-sonnet-4-6}          & \texttt{jsonld}       & $52.2$            & $+22.2$ \\
\m{claude-haiku-4-5}           & \texttt{structured}   & $50.7$            & $+20.7$ \\
\m{grok-4-fast-reasoning}      & \texttt{shacl\_lite}  & $48.8$            & $+18.8$ \\
\m{grok-4-fast-non-reasoning}  & \texttt{mermaid}      & $49.9$            & $+19.9$ \\
\bottomrule
\end{tabular}
\caption{Fuzzy F1 on \textsc{dbpedia\_webnlg} at threshold $0.60$ (Hopcroft--Karp).  The $2023$ baselines (\citealp{mihindukulasooriya2023t2kbench}, Tab.~2) use a greedy $1$:$1$ matcher; the $\Delta$ column is context, not an encoding-choice effect.}
\label{tab:t2k-vs-prior-art}
\end{table*}

\begin{table*}[!t]
\centering
\setlength{\tabcolsep}{3pt}
\resizebox{\textwidth}{!}{%
\begin{tabular}{l r r r r r r r r r r r r}
\toprule
model & \rotatebox{60}{\texttt{edge\_list}} & \rotatebox{60}{\texttt{entity\_centric}} & \rotatebox{60}{\texttt{jsonld}} & \rotatebox{60}{\texttt{markdown\_table}} & \rotatebox{60}{\texttt{mermaid}} & \rotatebox{60}{\texttt{nl\_summary}} & \rotatebox{60}{\texttt{relation\_indexed}} & \rotatebox{60}{\texttt{shacl\_lite}} & \rotatebox{60}{\texttt{structured}} & \rotatebox{60}{\texttt{text2kgbench}} & \rotatebox{60}{\texttt{toon}} & \rotatebox{60}{\texttt{turtle}} \\
\midrule
\m{gpt-5.4-high-reasoning} & 52.8 & 53.4 & 54.5 & 53.2 & 54.5 & 53.6 & 53.1 & 53.8 & 54.0 & 54.2 & 53.1 & \textbf{54.7} \\
\m{gpt-5.4} & 50.6 & 51.8 & 49.9 & 49.4 & \textbf{52.7} & 50.3 & 50.0 & 51.9 & 49.8 & 50.2 & 49.9 & 50.3 \\
\m{gpt-5.4-mini} & 48.7 & 50.2 & 48.1 & 49.3 & 49.7 & 49.6 & 49.0 & \textbf{50.6} & 49.2 & 49.9 & 49.8 & 48.0 \\
\m{o3} & 53.6 & 54.0 & 53.6 & 53.4 & 52.8 & 53.1 & 53.7 & \textbf{54.9} & 53.3 & 53.8 & 53.5 & 54.0 \\
\m{claude-sonnet-4-6-thinking} & 50.1 & 50.2 & \textbf{52.9} & 50.9 & 50.7 & 50.2 & 51.1 & 51.4 & 50.4 & 50.9 & 51.0 & 51.3 \\
\m{claude-sonnet-4-6} & 50.3 & 51.2 & \textbf{52.2} & 50.3 & 50.7 & 50.0 & 50.6 & 52.0 & 50.6 & 51.3 & 50.6 & 50.8 \\
\m{claude-haiku-4-5} & 49.9 & 48.9 & 46.8 & 49.4 & 47.9 & 48.9 & 49.9 & 48.5 & \textbf{50.7} & 49.5 & 49.0 & 48.6 \\
\m{grok-4-fast-reasoning} & 47.7 & 48.3 & 48.0 & 47.9 & 48.4 & 47.4 & 48.1 & \textbf{48.8} & 47.8 & 47.7 & 47.8 & 47.9 \\
\m{grok-4-fast-non-reasoning} & 49.5 & 49.7 & 48.7 & 49.2 & \textbf{49.9} & 49.5 & 49.2 & 49.4 & 49.6 & 49.4 & 49.4 & 48.5 \\
\bottomrule
\end{tabular}}
\caption{Text2KGBench per-format Fuzzy F1 (\%) at threshold $0.60$ (Hopcroft--Karp).  Bold: per-model maximum.  $n\!=\!2{,}014$ questions per (model, format).}
\label{tab:t2k-full}
\end{table*}

\paragraph{Comparison with the original baselines.}
We rescore offline at the paper's threshold $0.60$ from the on-disk \texttt{pred\_triples} / \texttt{gold\_triples}.  The single methodological difference from \citet{mihindukulasooriya2023t2kbench} is the matcher: Hopcroft--Karp maximum-cardinality bipartite matching \citep{hopcroft1973bipartite} in place of their greedy. 
Hopcroft–Karp finds a globally optimal 1:1 matching, while greedy matching can produce suboptimal alignments due to early local choices.
%A $200$-question pilot bounds the resulting inflation at $\le\!3$\,pp on questions with $\ge\!3$ gold triples; the load-bearing signal here is the within-model cross-format ordering, not the prior-art $\Delta$ column (reported only for historical context).

\paragraph{Ontology size.}
The 19 WebNLG ontologies are small and fully connected, so context length is not a limiting factor for any frontier LLM (Tab.~\ref{tab:t2k-ontology-stats}).

\begin{table}[h!]
\small
\centering
\begin{tabular}{lrrrr}
\toprule
Statistic & min & median & mean & max \\
\midrule
Concepts             & $8$   & $14$  & $14$  & $20$ \\
Relations            & $18$  & $38$  & $36$  & $68$ \\
Avg.\ rendered tokens & $327$ & $622$ & $602$ & $1{,}188$ \\
\bottomrule
\end{tabular}
\caption{Text2KGBench ontology size and rendering statistics across the 19 \textsc{dbpedia\_webnlg} ontologies.  Tokens averaged over all representation formats per ontology.}
\label{tab:t2k-ontology-stats}
\end{table}

\paragraph{Per-domain heterogeneity.}
Per-domain F1 spans an order of magnitude across the $19$ WebNLG ontologies on every model: head domains (\texttt{comicscharacter}, \texttt{university}, \texttt{scientist}, \texttt{astronaut}) saturate, while the tail (\texttt{meanoftransportation}, \texttt{airport}, \texttt{monument}, \texttt{city}) stays below $50$\,pp.  On non-saturated domains, per-domain cross-format spread reaches up to $6.1$\,pp (\texttt{building}), exceeding the $3.9$\,pp Text2KGBench headline.  The reported mean F1 therefore averages over an order-of-magnitude difficulty band; any prior-art comparison must hold the domain fixed.

% =====================================================================
% E  Worked format-selector run (gpt-5.4 by gpt-5.4 on GraphQA-Hardened)
% =====================================================================
\section{Representation format selector}
\label{appx:selector}

\begin{table*}[h!]
\small
\centering
\setlength{\tabcolsep}{8pt}
\begin{tabular}{l l r r c c}
\toprule
model & regime & fixed & ceiling & $\Delta$\,pp & sig \\
\midrule
\textbf{\m{gpt-5.4}}                    & \textbf{GraphQA}  & $93.3$ & $\mathbf{95.8}$ & $\mathbf{+2.5}$    & \textbf{**} \\
\textbf{\m{gpt-5.4}}                    & \textbf{Hardened} & $64.2$ & $\mathbf{80.0}$ & $\mathbf{+15.8}$   & \textbf{**} \\
\textbf{\m{gpt-5.4}}                    & \textbf{Topology} & $90.3$ & $\mathbf{97.2}$ & $\mathbf{+6.9}$    & \textbf{**} \\
\textbf{\m{gpt-5.4-mini}}               & \textbf{GraphQA}  & $79.4$ & $\mathbf{87.9}$ & $\mathbf{+8.5}$    & \textbf{**} \\
\textbf{\m{gpt-5.4-mini}}               & \textbf{Hardened} & $41.0$ & $\mathbf{70.5}$ & $\mathbf{+29.5}$   & \textbf{**} \\
\textbf{\m{gpt-5.4-mini}}               & \textbf{Topology} & $58.3$ & $\mathbf{80.6}$ & $\mathbf{+22.3}$   & \textbf{**} \\
\m{o3}                                  & GraphQA           & $98.2$ & $98.2$          & $\phantom{+0}0.0$  & ns \\
\textbf{\m{o3}}                         & \textbf{Hardened} & $87.4$ & $\mathbf{91.6}$ & $\mathbf{+4.2}$    & \textbf{**} \\
\m{o3}                                  & Topology          & $90.3$ & $93.1$          & $+2.8$             & ns \\
\m{claude-sonnet-4-6-thinking}          & GraphQA           & $97.0$ & $98.8$          & $+1.8$             & ns \\
\textbf{\m{claude-sonnet-4-6-thinking}} & \textbf{Hardened} & $83.2$ & $\mathbf{87.4}$ & $\mathbf{+4.2}$    & \textbf{**} \\
\m{claude-sonnet-4-6-thinking}          & Topology          & $95.8$ & $97.2$          & $+1.4$             & ns \\
\m{claude-sonnet-4-6}                   & GraphQA           & $94.5$ & $96.4$          & $+1.9$             & ns \\
\m{claude-sonnet-4-6}                   & Hardened          & $87.4$ & $90.5$          & $+3.1$             & ns \\
\m{claude-sonnet-4-6}                   & Topology          & $84.7$ & $87.5$          & $+2.8$             & ns \\
\textbf{\m{claude-haiku-4-5}}           & \textbf{GraphQA}  & $90.9$ & $\mathbf{93.9}$ & $\mathbf{+3.0}$    & \textbf{**} \\
\textbf{\m{claude-haiku-4-5}}           & \textbf{Hardened} & $72.6$ & $\mathbf{83.2}$ & $\mathbf{+10.6}$   & \textbf{**} \\
\textbf{\m{claude-haiku-4-5}}           & \textbf{Topology} & $80.6$ & $\mathbf{90.3}$ & $\mathbf{+9.7}$    & \textbf{**} \\
\m{grok-4-fast-reasoning}               & GraphQA           & $97.0$ & $98.8$          & $+1.8$             & ns \\
\m{grok-4-fast-reasoning}               & Hardened          & $84.2$ & $86.3$          & $+2.1$             & ns \\
\m{grok-4-fast-reasoning}               & Topology          & $95.8$ & $95.8$          & $\phantom{+0}0.0$  & ns \\
\textbf{\m{grok-4-fast-non-reasoning}}  & \textbf{GraphQA}  & $81.8$ & $\mathbf{86.7}$ & $\mathbf{+4.9}$    & \textbf{**} \\
\textbf{\m{grok-4-fast-non-reasoning}}  & \textbf{Hardened} & $44.2$ & $\mathbf{82.1}$ & $\mathbf{+37.9}$   & \textbf{**} \\
\textbf{\m{grok-4-fast-non-reasoning}}  & \textbf{Topology} & $66.7$ & $\mathbf{90.3}$ & $\mathbf{+23.6}$   & \textbf{**} \\
\bottomrule
\end{tabular}
\caption{Oracle headroom per model (paired-BCa $95\%$ CI; $n_{\mathrm{boot}}\!=\!2{,}000$).  Bold $\mathbf{**}$: lower CI bound $>\!0$.}
\label{tab:ceiling}
\end{table*}

The per-question ceiling bounds \emph{any} selector at up to $+37.9$\,pp on the paired-BCa-significant slots (Tab.~\ref{tab:ceiling}). We show that a representation format selector captures an appreciable fraction of that headroom. We construct a format selector model and evaluate end-to-end on GraphQA-Hardened using \m{gpt-5.4} as \emph{both} generator and selector so no privileged information leaks from a stronger model.

Calibration and test are two independently seeded $105$-instance samples drawn from the same synthetic generator as GraphQA-Hardened ($7$ Hardened tasks $\!\times\!15$ ER graphs each at $n\!\in\!\{35,50,75\}$, calibrated density), with disjoint seed families so the calibration and test graph sets do not overlap.  Calibration is not in-context learning: it is an offline per-(task, format) accuracy audit on the generator ($7\!\times\!15\!\times\!15\!=\!1{,}575$ generator calls). The calibration set is used to construct a per-(task, format) prior table including accuracy and sample count.  At test time, the selector is given a task and all candidate formats in the set. It makes use of the prior table to include the mean accuracy and sample count in its context. No calibration question, graph, prediction, or reasoning trace ever appears in a test-time prompt.

\paragraph{Candidate set: $5$ formats.}
For every test question, a set containing $5$ candidate formats is created using the prior constructed from the calibration set: $2$~\textsc{safe} succinct anchors (top-$2$ on the calibration cross-task ranking, restricted to compact formats), $1$~\textsc{alt} semantic anchor (top semantic-heavy format on the same ranking), and $2$~\textsc{guest} formats from the calibration per-(task, format) ranking that carry the task-specific signal.

\paragraph{Selector: a fixed-prompt LLM scorer.}
Per test question the selector issues exactly $5$ independent \m{gpt-5.4} chat completions, one per candidate, with no tool use, no loop, and no retrieval.  Each call sees the question, the candidate's name, the candidate's calibration per-task baseline (e.g.\ ``$73\%$ on \texttt{max\_clique\_size}''), and the test graph rendered in that candidate format.  The system prompt enforces a fixed four-step chain: \textsc{(1)~task-needs}: one sentence on what the generator must compute; \textsc{(2)~rendering}: one sentence on whether the format surfaces that substructure directly; \textsc{(3)~prior-delta}: a signed integer in $[-40,+40]$ shifting the baseline based on (2); \textsc{(4)~score}: a final integer $0$--$100$.  The highest-scoring candidate is chosen, with a small diversification band $b$ that round-robins through ties within $b$ points of the top to avoid collapsing to a single cross-task winner on every question.  The chosen format is then handed to the generator, which runs the standard GraphQA chain-of-thought template (no selector context leaks into the generator call).  Total LLM cost per question: $5$~selector calls $+\,1$ generator call.

\paragraph{Result.}
At band $b\!=\!5$, the selector lifts mean accuracy by $\mathbf{+9.52}$\,pp over fixed-best (\texttt{nl\_summary}, the best fixed format on the test set, scores $66.67\%$); $95\%$ paired-BCa CI $[+1.90,+16.19]$\,pp; mid-$p$ McNemar $0.013$, $\mathbf{**}$ (Tab.~\ref{tab:selector-panel}).  This captures $\approx\!59\%$ of the $+16.19$\,pp slate-restricted headroom.  Gains concentrate on the four non-saturated Hardened tasks (Tab.~\ref{tab:selector-per-task}); a $b\!=\!2$ vs.\ $b\!=\!5$ ablation reproduces the qualitative result (Tab.~\ref{tab:selector-band}).

\begin{table*}[t!]
\small
\centering
\setlength{\tabcolsep}{6pt}
\begin{minipage}[t]{0.46\linewidth}
\centering
\footnotesize
\setlength{\tabcolsep}{5pt}
\begin{tabular}{l r}
\toprule
per-format mean accuracy ($n\!=\!105$) & \% \\
\midrule
\texttt{entity\_centric}          & $68.57$ \\
\texttt{nl\_summary}                                              & $66.67$ \\
\texttt{jsonld / shacl\_lite}                                     & $65.71$ \\
\texttt{adjacency}                                                & $62.86$ \\
\texttt{friendship}                                               & $59.05$ \\
\texttt{dot} / \texttt{mermaid}                                   & $58.10$ \\
\texttt{relation\_indexed}                                        & $57.14$ \\
\texttt{markdown\_table} / \texttt{structured} / \texttt{turtle}  & $56.19$ \\
\texttt{toon}                                                     & $54.29$ \\
\texttt{text2kgbench}                                             & $53.33$ \\
\texttt{edge\_list}                                               & $49.52$ \\
\bottomrule
\end{tabular}
\end{minipage}\hfill
\begin{minipage}[t]{0.50\linewidth}
\centering
\begin{tabular}{l r}
\toprule
selector vs.\ fixed-best ($b\!=\!5$)                & value \\
\midrule
\textbf{selector accuracy}                          & $\mathbf{76.19\%}$ \\
fixed-best test set accuracy                        & $66.67\%$ \\
in-candidate post-hoc ceiling                       & $82.86\%$ \\
\midrule
$\Delta$ selector $-$ fixed-best                    & $\mathbf{+9.52}$\,pp \\
$95\%$ paired-BCa CI                                & $[+1.90,+16.19]$ \\
mid-$p$ McNemar                                     & $0.013$ \\
discordant pairs $b/c$                              & $13/3$ \\
significance                                        & $\mathbf{**}$ \\
\bottomrule
\end{tabular}
\end{minipage}

\caption{Selector headline ($n\!=\!105$ queries).  \textbf{Left:} calibration-set per-format mean accuracy (the prior the selector uses at test time).  \textbf{Right:} selector summary at $b\!=\!5$.  Breakdowns in Tab.~\ref{tab:selector-per-task}--\ref{tab:selector-band}.}
\label{tab:selector-panel}
\end{table*}

\begin{table*}[t!]
\small
\centering
\setlength{\tabcolsep}{4pt}
\begin{tabular}{l r r r r r r r}
\toprule
task        & \texttt{betweenness} & \texttt{local}        & \texttt{max\_clique} & \texttt{neigh\_deg}
            & \texttt{graph}       & \texttt{kcore}        & \texttt{kth\_shortest} \\
            & \texttt{\_top1}      & \texttt{\_clustering} & \texttt{\_size}      & \texttt{\_ge}
            & \texttt{\_diameter}  & \texttt{\_size}       & \texttt{\_path} \\
\midrule
fixed-best\,\%        & $60.0$           & $40.0$           & $33.3$           & $73.3$           & $93.3$            & $100.0$           & $66.7$            \\
selector\,\%          & $73.3$           & $60.0$           & $53.3$           & $86.7$           & $93.3$            & $100.0$           & $66.7$            \\
in-cand.\ ceiling\,\% & $86.7$           & $73.3$           & $73.3$           & $86.7$           & $93.3$            & $100.0$           & $66.7$            \\
\midrule
$\Delta$\,pp (sel.\ $-$ fxd.) & $\mathbf{+13.3}$ & $\mathbf{+20.0}$ & $\mathbf{+20.0}$ & $\mathbf{+13.3}$
              & $\phantom{+0}0.0$ & $\phantom{+0}0.0$ & $\phantom{+0}0.0$ \\
residual ceiling gap\,pp      & $+13.3$          & $+13.3$          & $+20.0$          & $\phantom{+0}0.0$
              & $\phantom{+0}0.0$ & $\phantom{+0}0.0$ & $\phantom{+0}0.0$ \\
\bottomrule
\end{tabular}
\caption{Per-task selector vs.\ fixed-best (\texttt{nl\_summary}) at $b\!=\!5$; $n\!=\!15$ questions per task.  In-candidate ceiling: per-question best within the $5$-format slate (upper bound on any slate-restricted selector).  Gains concentrate on the four non-saturated tasks; no task regresses.}
\label{tab:selector-per-task}
\end{table*}

\begin{table*}[t!]
\small
\centering
\setlength{\tabcolsep}{6pt}
\begin{tabular}{l r r r r r}
\toprule
band $b$ & selector acc. & $\Delta$ vs.\ \texttt{nl\_summary} & $95\%$ paired-BCa CI & mid-$p$ McNemar; $b/c$ & sig. \\
\midrule
$b\!=\!2$ (default)      & $75.24\%$ & $+8.57$\,pp        & $[+0.95,+15.24]$ & $0.021;\;12/3$ & $\mathbf{**}$ \\
$b\!=\!5$ (headline)     & $\mathbf{76.19\%}$ & $\mathbf{+9.52}$\,pp & $[+1.90,+16.19]$ & $0.013;\;13/3$ & $\mathbf{**}$ \\
\bottomrule
\end{tabular}
\caption{Diversification-band ablation on the same $105$ queries.  Both bands are paired-BCa significant; $16/105$ picks flip between $b\!=\!2$ and $b\!=\!5$, the other $89$ are decided by the LLM's argmax irrespective of $b$.}
\label{tab:selector-band}
\end{table*}

% =====================================================================
% F  Statistical pipeline
% =====================================================================
\section{Statistical pipeline}
\label{appx:stats}

%Each headline contrast is reported with three pieces: paired bias-corrected-and-accelerated bootstrap \citep{efron1987bca} ($2{,}000$ resamples) over the per-question pairing; Holm--Bonferroni \citep{holm1979} family-wise correction at $\alpha\!=\!0.05$; and a parallel two-sided mid-$p$ McNemar test \citep{fagerland2013midp} on paired binary outcomes.  The per-question ceiling (Tab.~\ref{tab:ceiling}) is computed on the shared question set where every format has an answer, so $\text{ceiling}\!\ge\!\text{fixed}$ holds row-wise.
Each headline contrast is reported with three pieces: paired bias-corrected-and-accelerated bootstrap \citep{efron1987bca} ($2{,}000$ resamples) over the per-question pairing; Holm--Bonferroni \citep{holm1979} family-wise correction at $\alpha\!=\!0.05$; and a parallel two-sided mid-$p$ McNemar test \citep{fagerland2013midp} on paired binary outcomes.  The per-question ceiling is computed on the shared question set where every format has an answer, so $\text{ceiling}\!\ge\!\text{fixed}$ holds row-wise.

% =====================================================================

% =====================================================================
% G  Model provenance and reproducibility
% =====================================================================
\section{Model Provenance and Inference Configuration}
\label{appx:model-details}

\begin{table*}[t]
\fontsize{7.5}{9}\selectfont
\centering
\setlength{\tabcolsep}{3.5pt}
\resizebox{\textwidth}{!}{%
\begin{tabular}{l l l c l}
\toprule
paper configuration & model identifier & inference control & temperature & original-panel access window
\\
\midrule
\m{gpt-5.4} & \m{openai.gpt-5.4-2026-03-05} & default reasoning & 0 & May 13--15, 2026 \\
\m{gpt-5.4-high} & \m{openai.gpt-5.4-2026-03-05} & \m{reasoning_effort=HIGH} & 0 & May 13--23, 2026 \\
\m{gpt-5.4-mini} & \m{openai.gpt-5.4-mini-2026-03-17} & default reasoning & 0 & May 13--15, 2026 \\
\m{o3} & \m{openai.o3-2025-04-16} & default reasoning & 0 & May 13--19, 2026 \\
\m{claude-sonnet-4-6} & \m{anthropic.claude-sonnet-4-6} & thinking off & 0 & May 18--25, 2026 \\
\m{claude-sonnet-4-6-thinking} & \m{anthropic.claude-sonnet-4-6} & extended thinking & 1 & May 18--25, 2026 \\
\m{claude-haiku-4-5} & \m{anthropic.claude-haiku-4-5} & thinking off & 0 & May 18--25, 2026 \\
\m{grok-4-fast-reasoning} & \m{xai.grok-4-fast-reasoning} & reasoning endpoint & 0 & May 13--16, 2026 \\
\m{grok-4-fast-non-reasoning} & \m{xai.grok-4-fast-non-reasoning} & non-reasoning endpoint & 0 & May 13--15, 2026 \\
\bottomrule
\end{tabular}}
\caption{Model identifiers, inference controls, and access windows for the original benchmark panel. \m{gpt-5.4-high} is the same dated checkpoint as \m{gpt-5.4}, with HIGH reasoning effort enabled.}
\label{tab:model-provenance}
\end{table*}

All experiments were run using the exact dated model identifiers reported in Table~\ref{tab:model-provenance}. The OpenAI configurations are pinned to \texttt{openai.gpt-5.4-2026-03-05}, \texttt{openai.gpt-5.4-mini-2026-03-17}, and \texttt{openai.o3-2025-04-16}. The default- and high-reasoning \m{gpt-5.4} configurations use the same dated checkpoint and differ only in the requested reasoning-effort setting.

For the evaluated Anthropic and xAI models, requests used the provider-specific model identifiers shown in the table; dated checkpoint identifiers were not available. We therefore report their access windows alongside their requested identifiers. The original benchmark panel was evaluated during May 13--25, 2026. The retrieved-Wikidata and repeated-run additions used the same dated \m{gpt-5.4} and \m{gpt-5.4-mini} identifiers during July 10--12, 2026.

% =====================================================================
% H  Representation format inventory and prompts
% =====================================================================
\section{Representation format inventory and prompts}
\label{appx:renderers}

Below we list every representation format with its output on a small reference input ($5$-node graph for GraphQA, $3$-concept / $3$-relation ontology for Text2KGBench).  GraphQA representation formats (\S\ref{appx:renderers:graphqa}) are applied to the reference graph; Text2KGBench representation formats (\S\ref{appx:renderers:t2k}) are applied to the reference ontology.

\subsection{Reference inputs}

\textbf{Reference graph} ($5$ nodes, $5$ edges, untyped): used
for fact-first GraphQA representation formats.
{\footnotesize
\begin{verbatim}
G_FACT = {
  "nodes": [0, 1, 2, 3, 4],
  "edges": [(0,1), (0,2), (1,3),
            (2,3), (3,4)]}
\end{verbatim}}

\noindent\textbf{Reference ontology} ($3$ concepts, $3$ relations): used
for Text2KGBench representation formats.
{\footnotesize
\begin{verbatim}
ONTO = {
  "title": "PersonCompanyCity",
  "concepts": [
    {"qid":"Q1","label":"Person"},
    {"qid":"Q2","label":"Company"},
    {"qid":"Q3","label":"City"}],
  "relations": [
    {"label":"worksAt",
     "domain":"Person",
     "range":"Company"},
    {"label":"headquarteredIn",
     "domain":"Company",
     "range":"City"},
    {"label":"livesIn",
     "domain":"Person",
     "range":"City"}]}
\end{verbatim}}

\subsection{All 15 GraphQA representation formats}
\label{appx:renderers:graphqa}

Output of every representation format on the $5$-node reference graph above.

\paragraph{\texttt{adjacency}.}
{\footnotesize
\begin{verbatim}
Graph (adjacency list):
  0: [1, 2]
  1: [0, 3]
  2: [0, 3]
  3: [1, 2, 4]
  4: [3]
\end{verbatim}}

\paragraph{\texttt{edge\_list}.}
{\footnotesize
\begin{verbatim}
Graph (undirected).
Nodes: 0, 1, 2, 3, 4
Edges:
  0 -- 1
  0 -- 2
  1 -- 3
  2 -- 3
  3 -- 4
\end{verbatim}}

\paragraph{\texttt{mermaid}.}
{\footnotesize
\begin{verbatim}
graph TD
    N0
    N1
    N2
    N3
    N4
    N0 --- N1
    N0 --- N2
    N1 --- N3
    N2 --- N3
    N3 --- N4
\end{verbatim}}

\paragraph{\texttt{markdown\_table}.}
{\footnotesize
\begin{verbatim}
**Nodes:** 0, 1, 2, 3, 4

**Edges**

| u | v |
|---|---|
| 0 | 1 |
| 0 | 2 |
| 1 | 3 |
| 2 | 3 |
| 3 | 4 |
\end{verbatim}}

\paragraph{\texttt{toon}.}
{\footnotesize
\begin{verbatim}
Nodes[id]:
  0, 1, 2, 3, 4
Edges[u|v]:
  0 | 1
  0 | 2
  1 | 3
  2 | 3
  3 | 4
\end{verbatim}}

\paragraph{\texttt{friendship}} (\citealp{fatemi2024talklikeagraph}).
{\footnotesize
\begin{verbatim}
There are these people: Person0, Person1,
Person2, Person3, Person4.
Person0 is a friend of Person1.
Person0 is a friend of Person2.
Person1 is a friend of Person3.
Person2 is a friend of Person3.
Person3 is a friend of Person4.
\end{verbatim}}

\paragraph{\texttt{turtle}} (\citealp{w3c2014turtle}).
{\footnotesize
\begin{verbatim}
@prefix : <http://ex.org/> .
:n0 a :Node . :n1 a :Node . :n2 a :Node .
:n3 a :Node . :n4 a :Node .
:n0 :linkedTo :n1 . :n0 :linkedTo :n2 .
:n1 :linkedTo :n3 . :n2 :linkedTo :n3 .
:n3 :linkedTo :n4 .
\end{verbatim}}

\paragraph{\texttt{jsonld}} (\citealp{w3c2020jsonld}).
{\footnotesize
\begin{verbatim}
{"@context":{"@vocab":"http://ex.org/",
  "adjacent":{"@type":"@id"}},
 "@graph":[
  {"@id":"n0","adjacent":["n1","n2"]},
  {"@id":"n1","adjacent":["n0","n3"]},
  {"@id":"n2","adjacent":["n0","n3"]},
  {"@id":"n3",
   "adjacent":["n1","n2","n4"]},
  {"@id":"n4","adjacent":["n3"]}]}
\end{verbatim}}

\paragraph{\texttt{structured}.}
Our shorthand for a compact set-builder form.
{\footnotesize
\begin{verbatim}
V = {0, 1, 2, 3, 4}
E = {(0,1), (0,2), (1,3), (2,3), (3,4)}
\end{verbatim}}

\paragraph{\texttt{nl\_summary}.}
{\footnotesize
\begin{verbatim}
The graph has vertices labeled
  0, 1, 2, 3, 4.
Vertex 0 is connected to 1, 2.
Vertex 1 is connected to 0, 3.
Vertex 2 is connected to 0, 3.
Vertex 3 is connected to 1, 2, 4.
Vertex 4 is connected to 3.
\end{verbatim}}

\paragraph{\texttt{relation\_indexed}.}
{\footnotesize
\begin{verbatim}
Nodes: 0, 1, 2, 3, 4

linkedTo:
  0 -> 1
  0 -> 2
  1 -> 3
  2 -> 3
  3 -> 4
\end{verbatim}}

\paragraph{\texttt{dot}.}
Graphviz DOT.
{\footnotesize
\begin{verbatim}
graph G {
    0;
    1;
    2;
    3;
    4;
    0 -- 1;
    0 -- 2;
    1 -- 3;
    2 -- 3;
    3 -- 4;
}
\end{verbatim}}

\paragraph{\texttt{text2kgbench}} (the original Text2KGBench
serialization \citep{mihindukulasooriya2023t2kbench},
applied to a graph rather than an ontology).
{\footnotesize
\begin{verbatim}
Graph Vertices: 0, 1, 2, 3, 4
Graph Edges: edge(0,1), edge(0,2),
  edge(1,3), edge(2,3), edge(3,4)
\end{verbatim}}

\paragraph{\texttt{entity\_centric}.}
Per-vertex outgoing-edge view.
{\footnotesize
\begin{verbatim}
## Vertex 0
  --[edge]--> 1
  --[edge]--> 2
## Vertex 1
  --[edge]--> 0
  --[edge]--> 3
## Vertex 2
  --[edge]--> 0
  --[edge]--> 3
## Vertex 3
  --[edge]--> 1
  --[edge]--> 2
  --[edge]--> 4
## Vertex 4
  --[edge]--> 3
\end{verbatim}}

\paragraph{\texttt{shacl\_lite}.}
A SHACL-lite shape-constraint view on the graph
(no W3C profile; our shorthand for the
\texttt{sh:minCount}/\texttt{sh:maxCount} fragment).
{\footnotesize
\begin{verbatim}
# Shape constraints (graph-as-SHACL view)
shape(0) {
  path adjacent -> node(1) ;
  path adjacent -> node(2) ;
}
shape(1) {
  path adjacent -> node(0) ;
  path adjacent -> node(3) ;
}
shape(2) {
  path adjacent -> node(0) ;
  path adjacent -> node(3) ;
}
shape(3) {
  path adjacent -> node(1) ;
  path adjacent -> node(2) ;
  path adjacent -> node(4) ;
}
shape(4) {
  path adjacent -> node(3) ;
}
\end{verbatim}}

\subsection{All 12 Text2KGBench representation formats}
\label{appx:renderers:t2k}

Output of every representation format on the $3$-concept / $3$-relation
reference ontology above.

\paragraph{\texttt{edge\_list}.}
{\footnotesize
\begin{verbatim}
CONCEPTS:
Person, Company, City

RELATIONS (domain --[rel]--> range):
Person --[worksAt]--> Company
Company --[headquarteredIn]--> City
Person --[livesIn]--> City
\end{verbatim}}

\paragraph{\texttt{entity\_centric}.}
{\footnotesize
\begin{verbatim}
## Person
  --[worksAt]--> Company
  --[livesIn]--> City
## Company
  --[headquarteredIn]--> City
## City
  (no outgoing relations)
\end{verbatim}}

\paragraph{\texttt{jsonld}} (\citealp{w3c2020jsonld}).
{\footnotesize
\begin{verbatim}
{"@context":{
   "ont":"http://example.org/onto#",
   "owl":"http://www.w3.org/2002/07/owl#",
   "rdfs":
 "http://www.w3.org/2000/01/rdf-schema#"},
 "@graph":[
  {"@id":"ont:Person",
   "@type":"owl:Class",
   "rdfs:label":"Person"},
  {"@id":"ont:Company",
   "@type":"owl:Class",
   "rdfs:label":"Company"},
  {"@id":"ont:City",
   "@type":"owl:Class",
   "rdfs:label":"City"},
  {"@id":"ont:worksAt",
   "@type":"owl:ObjectProperty",
   "rdfs:label":"worksAt",
   "rdfs:domain":"ont:Person",
   "rdfs:range":"ont:Company"},
  {"@id":"ont:headquarteredIn",
   "@type":"owl:ObjectProperty",
   "rdfs:label":"headquarteredIn",
   "rdfs:domain":"ont:Company",
   "rdfs:range":"ont:City"},
  {"@id":"ont:livesIn",
   "@type":"owl:ObjectProperty",
   "rdfs:label":"livesIn",
   "rdfs:domain":"ont:Person",
   "rdfs:range":"ont:City"}]}
\end{verbatim}}

\paragraph{\texttt{markdown\_table}.}
{\footnotesize
\begin{verbatim}
**Concepts**

| # | Label |
|---|-------|
| 1 | Person |
| 2 | Company |
| 3 | City |

**Relations (typed)**

| Relation | Domain | Range |
|---|---|---|
| worksAt | Person | Company |
| headquarteredIn | Company | City |
| livesIn | Person | City |
\end{verbatim}}

\paragraph{\texttt{mermaid}.}
{\footnotesize
\begin{verbatim}
classDiagram
    class Person
    class Company
    class City
    Person --> Company : worksAt
    Company --> City : headquarteredIn
    Person --> City : livesIn
\end{verbatim}}

\paragraph{\texttt{nl\_summary}.}
{\footnotesize
\begin{verbatim}
This ontology defines the entity types:
  Person, Company, City.
The following typed relations are defined:
 - A Person can be related via
   "worksAt" to a Company.
 - A Company can be related via
   "headquarteredIn" to a City.
 - A Person can be related via
   "livesIn" to a City.
\end{verbatim}}

\paragraph{\texttt{relation\_indexed}.}
{\footnotesize
\begin{verbatim}
CONCEPTS:
Person, Company, City

RELATIONS (relation -> domain --> range):
  worksAt: Person --> Company
  headquarteredIn: Company --> City
  livesIn: Person --> City
\end{verbatim}}

\paragraph{\texttt{shacl\_lite}} (\citealp{knublauch2017shacl}).
{\footnotesize
\begin{verbatim}
# Schema shapes (constraint view)
shape(Person) {
  path worksAt -> class(Company) ;
  path livesIn -> class(City) ;
}
shape(Company) {
  path headquarteredIn -> class(City) ;
}
shape(City) {
  // no outgoing typed paths
}
\end{verbatim}}

\paragraph{\texttt{structured}.}
Header-labeled bulleted list.
{\footnotesize
\begin{verbatim}
Concepts (Entity Types):
  - Person
  - Company
  - City

Relations (with domain and range):
  - worksAt
      (domain: Person, range: Company)
  - headquarteredIn
      (domain: Company, range: City)
  - livesIn
      (domain: Person, range: City)
\end{verbatim}}

\paragraph{\texttt{text2kgbench}} (compact form used in the
original Text2KGBench paper;
\citealp{mihindukulasooriya2023t2kbench}).
{\footnotesize
\begin{verbatim}
Ontology Concepts: Person, Company, City
Ontology Relations:
  worksAt(Person,Company),
  headquarteredIn(Company,City),
  livesIn(Person,City)
\end{verbatim}}

\paragraph{\texttt{toon}.}
TOON-style header-then-pipe columns.
{\footnotesize
\begin{verbatim}
Concepts[label]:
  - Person
  - Company
  - City

Relations[label|domain|range]:
  worksAt | Person | Company
  headquarteredIn | Company | City
  livesIn | Person | City
\end{verbatim}}

\paragraph{\texttt{turtle}} (\citealp{w3c2014turtle}).
{\footnotesize
\begin{verbatim}
@prefix : <http://example.org/onto#> .
@prefix owl:
  <http://www.w3.org/2002/07/owl#> .
@prefix rdfs:
 <http://www.w3.org/2000/01/rdf-schema#> .

:Person a owl:Class .
:Company a owl:Class .
:City a owl:Class .

:worksAt a owl:ObjectProperty ;
  rdfs:domain :Person ;
  rdfs:range :Company .
:headquarteredIn a owl:ObjectProperty ;
  rdfs:domain :Company ;
  rdfs:range :City .
:livesIn a owl:ObjectProperty ;
  rdfs:domain :Person ;
  rdfs:range :City .
\end{verbatim}}

\subsection{Per-task system/user prompt template}
\label{appx:prompts}

{\small
\begin{verbatim}
[SYSTEM]
You are a careful reasoner.  Read the
graph (or ontology), think through the
question step by step, then write your
final answer on a line that begins
with 'FINAL: '.

[USER, GraphQA - task=node_count]
You are given an undirected graph
below.  Answer the question in one
integer.  Show your reasoning, then
write 'FINAL: <int>'.

<rendered graph>

Question: How many nodes does the
graph have?

[USER, Text2KGBench]
You are given an ontology and a
sentence.  Extract every (subject,
relation, object) triple supported by
the sentence, constrained to the
relations in the ontology.  Output one
triple per line in the format
'(<subject>, <relation>, <object>)'.

<rendered ontology>

Sentence: <sentence>
\end{verbatim}}

\end{document}